\def\year{2020}\relax
\documentclass[letterpaper]{article} 
\usepackage{aaai20}  
\usepackage{helvet} 
\usepackage{courier}  
\usepackage[hyphens]{url}  
\usepackage{graphicx} 
\usepackage{graphicx}  
\usepackage{amsmath}
\usepackage{amssymb}
\usepackage{multirow}
\usepackage[linesnumbered,ruled,vlined]{algorithm2e}
\usepackage{algpseudocode}
\title{Domain Adaptive Attention Learning for Unsupervised Person Re-Identification}
\author{Yangru Huang\textsuperscript{\rm 1}\footnotemark[1],
Peixi Peng\textsuperscript{\rm 2}\thanks{Equal contribution.},
Yi Jin\textsuperscript{\rm 1}\thanks{Corresponding author.},
Yidong Li\textsuperscript{\rm 1},
Junliang Xing\textsuperscript{\rm 2},
Shiming Ge\textsuperscript{\rm 3} \\ 
\textsuperscript{\rm 1} School of Computer and Information Technology, Beijing Jiaotong University, China\\
\textsuperscript{\rm 2} Institute of Automation, Chinese Academy of Sciences, China\\
\textsuperscript{\rm 3} Institute of Information Engineering, Chinese Academy of Sciences, China\\
\small {yrhuang@bjtu.edu.cn, {peixi.peng}@ia.ac.cn, yjin@bjtu.edu.cn }\\
\small {ydli@bjtu.edu.cn, jlxing@nlpr.ia.ac.cn, geshiming@iie.ac.cn}\\
}
 
\begin{document}

\maketitle

\begin{abstract}
Person re-identification (Re-ID) across multiple datasets is a challenging task due to two main reasons: the presence of large cross-dataset distinctions and the absence of annotated target instances. To address these two issues, this paper proposes a domain adaptive attention learning approach to reliably transfer discriminative representation from the labeled source domain to the unlabeled target domain. In this approach, a domain adaptive attention model is learned to separate the feature map into domain-shared part and domain-specific part. In this manner, the domain-shared part is used to capture transferable cues that can compensate cross-dataset distinctions and give positive contributions to the target task, while the domain-specific part aims to model the noisy information to avoid the negative transfer caused by domain diversity. A soft label loss is further employed to take full use of unlabeled target data by estimating pseudo labels. Extensive experiments on the Market-1501, DukeMTMC-reID and MSMT17 benchmarks demonstrate the proposed approach outperforms the state-of-the-arts.
\end{abstract}

\section{Introduction}

The task of person re-identification (Re-ID) is to match people across non-overlapping camera views. It has become
one of the most studied problems in video surveillance
due to its great potential for security and safety management
applications.
It is a challenging task because a person's appearance often
changes dramatically across camera views caused by changes
in body pose, view angle, occlusion and illumination condition.

\begin{figure}[t]
\centering
\includegraphics[width=\columnwidth]{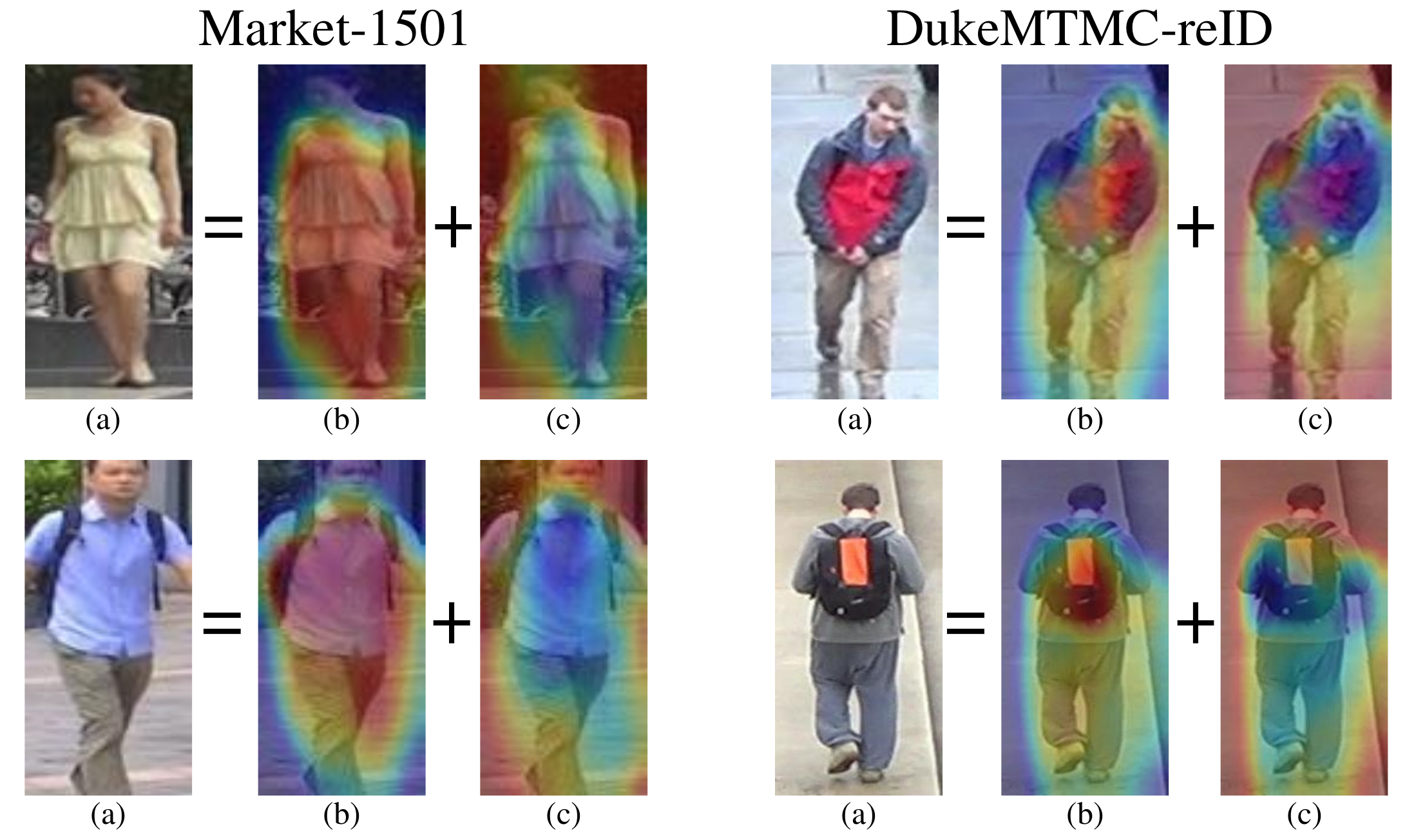}
\caption{ The visualizations of the proposed DAAM. The DAAM (the warmer color means the greater weight) separates the original image (a) into two parts: the domain-shared part (b) and the domain-specific part (c). The former is discriminative and useful to the person Re-ID task, and the latter is used to model the domain-specific information (such as background) caused by the domain divergence.  }
\label{fig:attention_v}
\end{figure}

In order to address these issues, most of the existing person Re-ID methods are designed on supervised learning~\cite{sun2018beyond,wang2018learning,reid} and have obtained significant performance improvement by the deep learning~\cite{Densenet,Chollet2017Xception,He2016DeepRL}. However, these methods require a large number of labeled data to train the Re-ID model, which are limited in many real-world applications \cite{Peng2016UnsupervisedCT}. In order to make person Re-ID method more scalable, one solution is to formulate the person Re-ID task as an unsupervised domain adaption problem (UDA) \cite{fernando2013unsupervised,gong2012geodesic,gopalan2011domain,long2016unsupervised}, where the existing labeled dataset and the current unlabeled dataset are modeled as source and target domains, respectively.  The source and target domains contain the identical feature space with the same dimension but totally different person identities (IDs). It is a challenging task to transfer a Re-ID model from the source domain to the target domain due to two reasons: Firstly,
since the source and target datasets are often collected from totally different environments which contains various illuminations, backgrounds and image qualities, the data distributions of the source and target data are different with a large probability and the domain divergence \cite{ben2007analysis,ben2010theory} may cause negative transfer.
Secondly, the target dataset is unlabeled, while most of the existing loss functions of Re-ID are designed for the supervised learning and cannot be employed directly.

To handle the first challenge, a novel domain adaptive attention module (DAAM) is proposed to alleviate the negative transfer caused by the domain divergence. Given a feature map of any image from a backbone network, the proposed domain adaptive attention module (DAAM) aims to separate the feature map into the domain-shared (DSH) feature map and the domain-specific (DSP) feature map simultaneously. Specifically, the DAAM focuses on capturing the attentive parts of the DSH feature map which is discriminative and transferable, and thus it can help the Re-ID task in the target dataset. Then, the DSP feature map is used to model the residual part corresponding to the domain specific information such as background. The residual mechanism encourages the DSH and DSP feature maps separable and complementary to each other.
Although the DSP feature map is useless for the Re-ID task, it makes sure that the domain-specific information is accounted for in the model rather than acting as a distracter to corrupt the learning of the DSH feature map. Several visualized examples of the DAAM are shown in Fig.~\ref{fig:attention_v}. Following the DSH and DSP feature map, two branches are introduced respectively. Then, the DSH branch is trained by the person Re-ID loss to ensure the DSH feature map discriminatively, and a domain-specific loss is introduced to ensure that the DSP feature map is distinguishable for different domains.

To take full use of the unlabeled target data, the pseudo labels~\cite{fan2018unsupervised,song2018unsupervised,lin2019bottom,yu2019unsupervised,SSG,zhong2019invariance} are widely used to the unsupervised Re-ID task. The standard pipeline often employs the clustering methods (such as DBSCAN~\cite{ester1996density}) to segment the unlabeled training data into several independent clusters, and assumes that the data in the same cluster have same person ID.
However, different with manually annotation, the pseudo labels are approximated and inaccurate. Hence, we consider the pseudo labels as the soft constrains, and a novel soft person Re-ID loss is proposed according to the relationship between the training data and the clusters. Specifically, the clusters are regarded as the potential IDs, and the pseudo labels are assigned as possibility distributions rather than definitely ID~\cite{fan2018unsupervised,song2018unsupervised}.

\begin{figure*}
\begin{center}
\includegraphics[width=\textwidth]{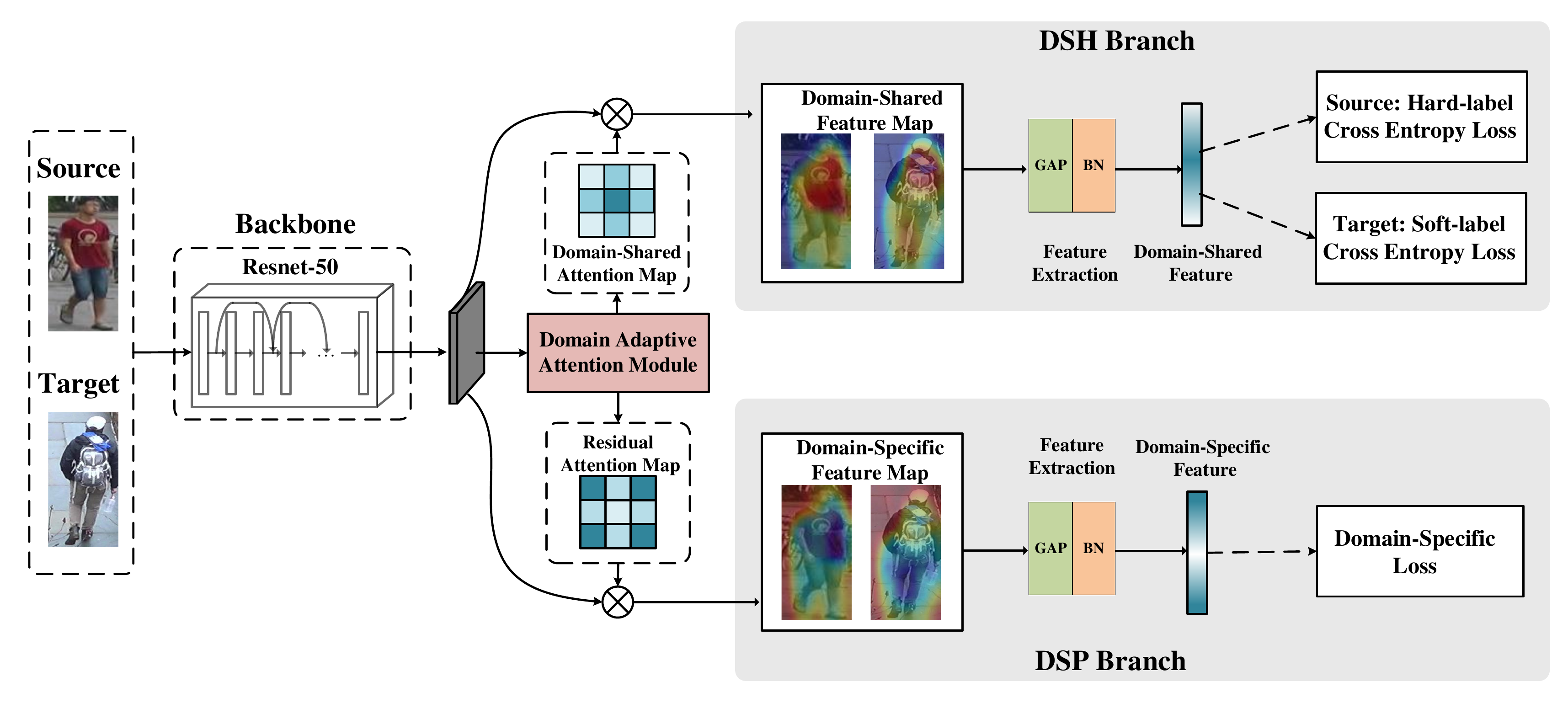}
\end{center}
   \caption{The framework of the proposed method. The domain adaptive attention module separates the feature map into domain-shared (DSH) and domain-specific (DSP) feature maps simultaneously. Then, a DSH branch  and a DSP branch are introduced to learn these two feature maps respectively. The soft-label based cross entropy and hard-label based losses are adopted to the DSH branch to make the DSH feature discriminative to different persons and transferable to different domains. In contrary, the domain-specific loss is employed to make the DSP branch capture the domain distinguishable information to avoid to distract the DSH feature and separate the untransferable to different domains.
}
\label{fig:2}
\end{figure*}

The whole framework of the proposed method is shown in Fig.~\ref{fig:2}, and the main technical contributions are outlined as below:
\begin{itemize}
\item A novel domain adaptive attention model is proposed to automatically separate the feature map of an image to the domain-shared feature map and the domain-specific feature map simultaneously.
\item A soft label based person Re-ID loss is introduced for the unlabeled target dataset.
\item Extensive experimental analyses and evaluations on the Market-1501, DukeMTMC-reID and MSMT17 benchmarks demonstrate the proposed method can achieve the state-of-the-art performance.
\end{itemize}

\section{Related Works}

Person Re-ID has been one of the most studied problems due to its important application, and most of the existing works are based on supervised learning frameworks \cite{sun2018beyond,wang2018learning,reid} which require sufficient labeled images across cameras. This severely limits the scalability of these supervised learning based methods.%
To solve the above scalability issue, a natural solution is to utilize unsupervised domain adaption method which aims to transfer useful Re-ID information from the labeled source domain (dataset) to unlabeled target domain (dataset). However, most of existing unsupervised domain adaptation methods \cite{long2016unsupervised,DSN,tzeng2017adversarial} are designed to the cases where the source and target domains have the same recognition tasks (\textit{i.e.} having the same set of classes), which is invalid to person Re-ID problem as different datasets contain totally different person identities.
In order to address the problem, early unsupervised domain adaptation methods ~\cite{ma2015cross,Peng2016UnsupervisedCT}
are proposed based on hand-craft features, but they are less effective than the deep model when a large number of training samples are available. Recently, several deep-learning-based methods are proposed and mainly can be categorised into two groups:

\noindent\textbf{Pseudo Label Based Methods.} Another focus of unsupervised cross-domain person Re-ID works is estimating pseudo identity labels~\cite{fan2018unsupervised,song2018unsupervised,lin2019bottom,yu2019unsupervised,SSG,zhong2019invariance} for target dataset which is similar to our method. 
PUL~\cite{fan2018unsupervised}, BUC~\cite{lin2019bottom}, SSG~\cite{SSG} and UDAP~\cite{song2018unsupervised} generate hard pseudo labels for unlabeled target data by iterative clustering. The key different with them, in view of inaccuracy of pseudo labels, our method estimates probabilistic class labels for unlabeled samples in a soft way like LSR~\cite{szegedy2016rethinking}.
However, LSR assigns a uniform label distribution to all samples, while our method assigns a label distribution according to the reliability of pseudo label.
ECN~\cite{zhong2019invariance} uses k-nearest neighbors to exploit the target labels also in a soft way, but it introduces the camera-invariance into the model,which requires that each real image and its style-transferred counterparts share the same identity.
MAR~\cite{yu2019unsupervised} using soft multilabel representing an unlabeled target person by other different reference persons, while our soft labels not only used to represent an unlabeled target person but also measure the reliability of the pseudo labels.

\noindent\textbf{GAN Based Methods.} To reduce the discrepancy, several methods employ GAN to transfer the style of different domain person images \cite{Wei2017PersonTG,Deng2017ImageImageDA,zhong2018camera,liu2018pose,zhong2019invariance,qi2019novel}. The proposed method does not contradict with those GAN based methods. They aim to increase the cross-domain training samples with style transfer methods, while our goal is to make better use of existing samples. Furthermore, their generated images can also be used in our method as training samples.

Besides, there are other deep transfer learning approaches~\cite{wang2018transferable,huang2018eanet,li2018adaptation}. Wang \textit{et al.} \cite{wang2018transferable} introduce a transferable joint attribute-identity deep learning to the target domain for Re-ID tasks, but the method requires additional attribute annotations. EANet~\cite{huang2018eanet} also introduces additional pose segmentation infromation to enhance alignment. However, our method does not require any additional information.
ARN \cite{li2018adaptation} extends the domain separation network \cite{DSN} for person Re-ID task which leverages encoder to model domain-shared and domain-specific features.
Different with these two methods, we propose a domain adaptive attention module (DAAM) to separate feature map which is more direct than the learned encoder.
Based on the above analysis, in this paper, we aim to address unsupervised domain adaptation problem for person Re-ID by learning domain adaptive attention representations based on soft labels.

\section{Methodology}
Suppose there are two types of training data including:
a labeled source dataset $ D_s=\{x_i^s,y_i^s\}_{i=1}^{N_s} $ and an unlabeled target dataset $ D_t=\{x_i^t\}_{i=1}^{N_t}$, where  $x_i^s$ and $x_i^t$ are pedestrian images from the source and target datasets respectively. The person identity (ID) of each image is only available in the source training data and denoted by $y_i^s\in \{1,2,...,N^{ids}\}$.
The source and target datasets are collected from different environments and have different data distributions. To transfer the dataset-shared discriminative representations from the source dataset to the target dataset (the source and target datasets are represented by source and target domains respectively), person Re-ID task is formulated as an unsupervised domain adaptation problem \cite{long2016unsupervised}, where the labeled source and unlabeled target domains contain the identical feature space with the same dimension but totally different IDs. To resolve this problem, as is shown in Fig.~\ref{fig:2}, a novel deep network is designed which consists of four modules: a backbone network, a domain adaptive attention module, a domain-shared branch and a domain-specific branch. ResNet-50 is chosen as our backbone network as same to most recent person Re-ID methods \cite{wang2018transferable,li2018adaptation,fan2018unsupervised}. Given any image $x$, the output of the backbone network is the corresponding feature map  $F_x \in \mathcal{R}^{h\times w\times c}$ where $h$, $w$ and $c$ are the height, width and the number of channels, respectively.

\subsection{Domain Adaptive Attention Module}
Given $F_x$, the goal of domain adaptive attention module (DAAM)  is to focus on the domain shared (DSH) and discriminative part of $F_x$ while eliminating the irrelevant domain specific (DSP) noise such as backgrounds.  Specifically, the input of DAAM is $F_x$, while the output is the DSH attention maps $A(F_x)\in (0,1)^{h\times w\times c}$. Then, the DSH feature map can be calculated as:
\begin{align}
\small
\begin{aligned}
\label{eq:attention}
F_x^{sh}=A(F_x)\otimes F_x,
\end{aligned}
\end{align}
where $\otimes$ means element-wise product. Due to the irrelevant and complementary of the DSH feature map $F_x^{sh}$ and the DSP feature map $F_x^{sp}$, we have:
\begin{align}
\small
\begin{aligned}
\label{eq:attention1}
F_x^{sp}=(1-A(F_x))\otimes F_x.
\end{aligned}
\end{align}

\begin{figure}[t]
\centering
\includegraphics[width=\columnwidth]{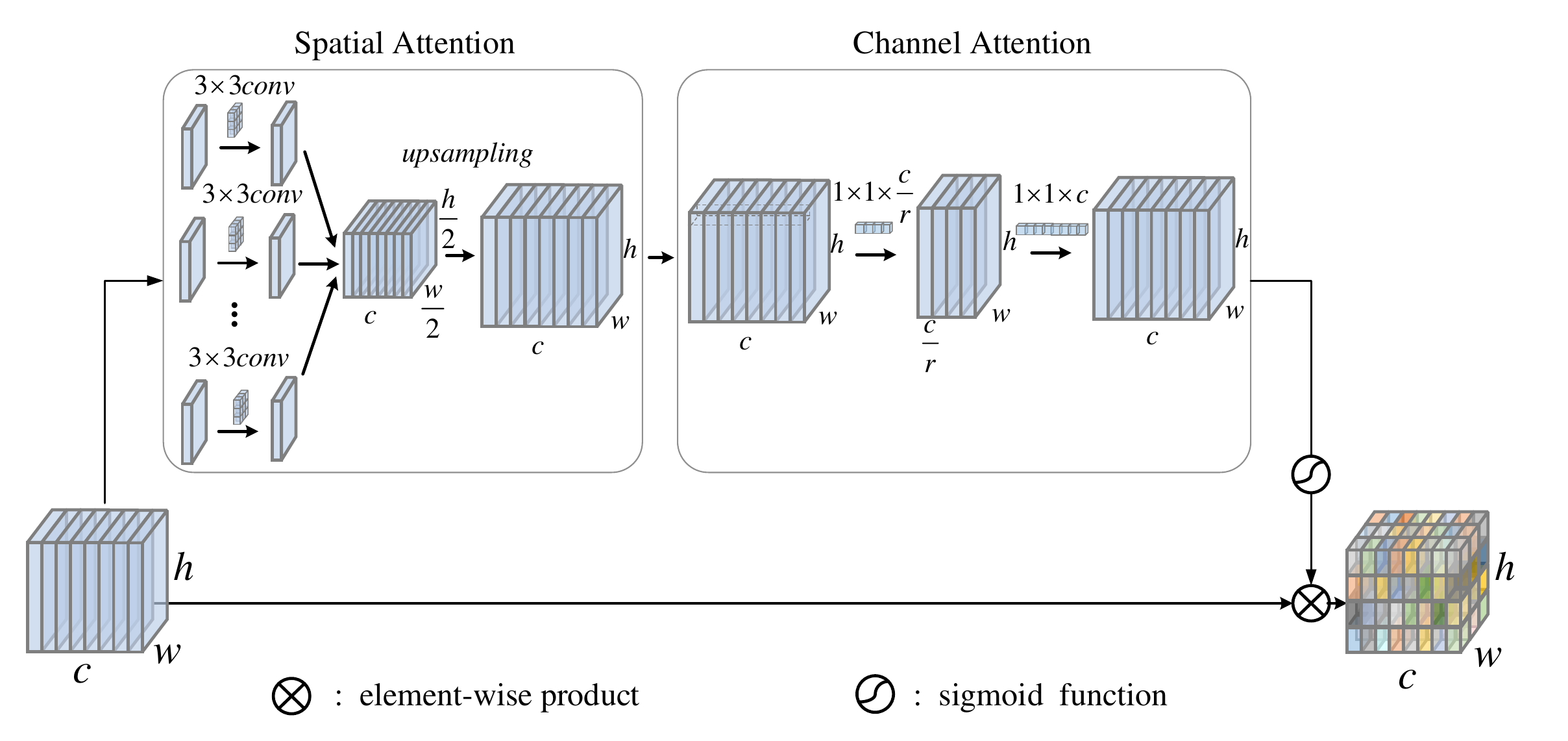}
\caption{~~Diagram of domain adaptive attention module. 
}
\label{fig:attention}
\end{figure}

Inspired by many recent attention techniques \cite{reid,chen2019abdnet}, we learn the spatial attention and the channel attention sequentially. For the spatial attention, a depthwise separable convolution layer containing $\frac{c}{2}$ $3\times 3 $ kernels are employed on $F_x$ where the stride is 2. The layer exploits the inter-pixel relationship of feature maps and preserves channel-specific characteristics. After that, a resizing bilinear layer is introduced to make the attention map have same size with $F_x$. For the channel attention, two convolution layers with $1\times 1$ kernels are introduced to exploit the inter-channel relationship of feature maps. Specifically, the first layer contains $\frac{c}{16}$ kernels and project the feature map to a new down sampling channel space, and then the second layer with $c$ kernels is performed to recover the size.  In addition, the batch normalization layer and ReLU activation function are followed by each above convolution layer, and the network architecture of the DAAM is shown in Fig.~\ref{fig:attention}.

\subsection{Domain-Shared Branch}
The domain-shared (DSH) branch is designed to extract feature representations from $F^{sh}_x$ which are applied for the person Re-ID task at the target domain.  Specifically, a Global Average Pooling (GAP) operation is performed to get the feature vector $f_x^{sh}$.
To make $f_x^{sh}$  discriminative to different persons, two types of person Re-ID loss functions, including the hard-label based cross entropy loss and soft-label based cross entropy loss, are introduced for source dataset and target respectively.

\noindent\textbf{Hard-label Based Cross Entropy Loss.}
For the labeled source dataset, the cross entropy loss is adopted as same to the existing supervised learning methods. Considering every person as a class, and person Re-ID task is formulated as a classification task. That is, the label of each sample is hard and denoted by one-hot coding.
Therefore, a $2048\times N^{ids}$ FC layer and a softmax activate function are performed sequentially following $f^{sh}_x$ to output the probabilities $\{p^{id}(y|x)\}_{y=1}^{N^{ids}}$ of that the image $x$ is from person $y$. Then, the cross-entropy loss is formulated as:
\begin{equation}
\begin{aligned}
\mathcal{L}^{cro_s}=-\log(p^{id}(y^s_i|x^s_i)).
\label{eq:cro_s}
\end{aligned}
\end{equation}

\noindent\textbf{Soft-label Based Cross Entropy Loss.}
Since the training data in the target domain are unlabeled, Eq.~(\ref{eq:cro_s}) cannot be directly applied to the target dataset. Motivated by \cite{Peng2016UnsupervisedCT},  we assume that images with more similar appearances are more likely from the same person. Based on this assumption, a widely used solution \cite{fan2018unsupervised,song2018unsupervised,SSG} is to estimate the pseudo label of unlabeled data by the clustering method. Specifically, the feature $f^{sh}_x$ of the target data is extracted by the pre-trained model, and the distance matrix $D$ between samples can be calculated by the k-reciprocal encoding~\cite{Zhong_2017_CVPR}. As same to \cite{song2018unsupervised,SSG}, the density-based clustering method DBSCAN \cite{ester1996density} is used to segment the train data
of the target domain into $K$ groups, denoted as $\{\mathcal{C}_k\}_{k=1}^K$ respectively. Then, most existing methods \cite{fan2018unsupervised,song2018unsupervised,SSG} assign the pseudo label $\tilde y^t_i$ of a sample $x^t_i$ by which group $\mathcal{C}_k$
is belongs. That is, if $f^{sh}_{x^t_i} \in \mathcal{C}_k$, $\tilde y^t_i=k$. Based on $\tilde y^t_i$, the supervised losses are directly calculated.

However, unlike manually annotations, $\tilde y^t_i$ is approximated and may be inaccurate. Motivated by LSR~\cite{szegedy2016rethinking}, if the model learns to assign full probability to the pseudo label for each training example, it is easy to get caught up in overfitting on unreliable pseudo labels.
To handle this challenge, we argue that the model should have less confident on the approximated pseudo labels and mine the potential association information between different groups. Following this line, a soft label based loss is proposed. Different with the hard label, the soft label is denoted by $K$ weights which is used to measure the relationship between the sample with $K$ groups.
Generally speaking, the sample closer to a group center $\mathcal{C}_k$ should have a larger confidence, hence the weight of sample belonging to each center $\{w_{i,k}\}^K_{k=1}$ is defined as a descending function of the distance between $x^t_i$ and every group $\{\mathcal{C}_k\}_{k=1}^K$:
\begin{equation}
\label{eq:weight}
w_{i,k}=\left\{\begin{matrix}
\frac{1}{K \times {\|f^{sh}_{x^t_i}-\mathcal{C}_{k}\|^2_2} }, & k\neq \tilde y^t_i  & \\
 \epsilon+\frac{1 }{K \times{\|f^{sh}_{x^t_i}-\mathcal{C}_{k}\|^2_2}},& k= \tilde y^t_i,& \\
\end{matrix}\right.
\end{equation}
where $k \in \{1,2,...,K\}$ and $\epsilon \in [0,1]$. $\epsilon$ has the function of ensuring a minimum weight for the assigned cluster and encourages the weights of the group $\mathcal{C}_k$ containing $x_i^t$ when $k= \tilde y^t_i$, and is set to 0.9 in our experiments.

Combining the soft label and the cross entropy loss, the person Re-ID loss of the unlabel target data is formulated as:
\begin{equation}
\begin{aligned}
\mathcal{L}^{cro_t}=-\sum_{k = 1}^{K}w_{i,k} \log(p^{ {id}}( k|x^t_i)).
\label{eq:cro_t}
\end{aligned}
\end{equation}
where $\{p^{id}(k|x)\}_{k=1}^K$ is calculated by performing a $2048\times K$ FC layer and a softmax function sequentially to  $f^{sh}_{x}$.

\subsection{Domain-Specific Branch}
To make the model easy to learn, the domain-specific attention map is approximated by Eq.~(\ref{eq:attention}). Similar to the DSH branch, the domain-specific (DSP) branch adopts a GAP operation to get a 2048-dimensional feature vector $f_x^{sp}$.
And the network parameters of the DSP branch are independent to DSH branch since they are designed to different tasks. To ensure $f_x^{sp}$ domain-specific, $f_x^{sp}$ should be distinguishable to different domains. Therefore, a domain classifier is introduced which is composed of a $2048\times2$ FC layer and a softmax function, and it is used to predict the probabilities ($p^s(x)$ and $p^t(x)$) of that image $x$ is from the source domain or the target domain, respectively. The domain classifier should predict the domain well, and the domain-specific loss is defined as a cross-entropy loss:
\begin{equation}
\begin{split}
\mathcal{L}_{DSP}=\left\{\begin{matrix}
-\log(p^s(x^s_i)), & x_i \in D_s  & \\
-\log(p^t(x^t_i)), & x_i \in D_t. & \\
\end{matrix}\right.
\label{eq:specficloss}
\end{split}
\end{equation}

Finally, the proposed network is optimized by minimizing Eq.~(\ref{eq:cro_s}), Eq.~(\ref{eq:cro_t}) and Eq.~(\ref{eq:specficloss}) jointly. The total loss function is defined as:
\begin{equation}
\begin{aligned}
\label{eq:total}
\mathcal{L}_{total}=\left\{\begin{matrix}
\mathcal{L}^{cro_s}+\mathcal{L}^{DSP}, & x_i \in D_s  & \\
\mathcal{L}^{cro_t}+\mathcal{L}^{DSP}, & x_i \in D_t & \\
\end{matrix}\right.
\end{aligned}
\end{equation}

\section{Learning}

In the learning procedure, we firstly pre-train the network on the labeled source dataset in a supervised way only using $\mathcal{L}^{cro_s}$. Then, $\{f^{sh}_x\}_{x\in D_t}$ of target data are extracted by the pre-trained model, and the pseudo labels $\{\tilde y_x\}_{x\in D_t}$ of the target data are estimated by performing the clustering method DBSCAN to the distance matrix of $\{f^{sh}_x\}_{x\in D_t}$ after the k-reciprocal encoding, and sample weights in Eq.~(\ref{eq:cro_t}) is calculated by Eq.~(\ref{eq:weight}) with $\{f^{sh}_x\}_{x\in D_t}$. Note that, DBSCAN has two hyper-parameters: $eps$ and the minimum number $M$ of points required to form a dense region.
In order to estimate $eps$ effectively, we firstly calculate and sort the distance between all target samples, and then select a certain proportion $p$ to get the average distance as $eps$. Secondly, the whole network is updated by minimizing $\mathcal{L}_{total}$ over both source and target datasets, and new $\{f^{sh}_x\}_{x\in D_t}$ are extracted. Then, the pseudo labels and sample weights are updated by the new $\{f^{sh}_x\}_{x\in D_t}$, and we re-train the network by updated pseudo labels and sample weights to enter the next iteration. The iterations terminate when
a stopping criterion is met, and the number of iterations is typically $<$ 10 in our experiments. Alg.~\ref{alg:1} concludes the proposed learning method.

\begin{algorithm}
  \label{alg:1}
  \caption{The proposed learning algorithm.}
  \KwIn{Labeled source data $D_s=\{x_i^s,y_i^s\}_{i=1}^{N_s}$, unlabeled target data $D_t=\{x_i^t\}_{i=1}^{N_t}$, ~~~~percentage $p$, the mininum size of a cluster $M$, iteration number $Iteration$;\\
  }
  \KwOut{ The trained network parameters.}
  Pre-train the network on $D_s$ by $\mathcal{L}^{cro_s}$.\\
  \For{$iter=1,...,Iteration$}
  {
  Extract  $\{f^{sh}_x\}_{x\in D_t}$.\\
  Update $\{\tilde y_i\}_{i=1}^{N_t}$ and $\{\mathcal{C}_k\}_{k=1}^K$ by DBSCAN.\\
  Compute sample weights according to Eq.~(\ref{eq:weight}).\\
  Update the network by minimizing  Eq.~(\ref{eq:total}).\\
  }
\end{algorithm}
\section{Experiment}
\subsection{Datasets}
\textbf{Market-1501}~\cite{zheng2015scalable} contains 32, 668 images of 1, 501 identities captured by 6 camera views. The pedestrians are cropped with bounding-boxes predicted by DPM detector \cite{DPM}. Following the standard setting \cite{zheng2015scalable}, the whole dataset is divided into a training set containing 12, 936 images of 751 identities and a testing set containing 19, 732 images of 750 identities.
\textbf{DukeMTMC-reID}~\cite{ristani2016performance} consists of 36,411 images of 1,812 persons from 8 high-resolution cameras, where 1,404 people appear more than two cameras and other 408 people images are regarded as distractors. 16,522 images of 702 persons are randomly selected from the dataset as the training set, and the remaining 702 persons are divided into the
testing set where contains 2,228 query images and 17,661 gallery images. The dataset split setting is same to \cite{ristani2016performance}.
\textbf{MSMT17}~\cite{Wei2017PersonTG} is a larger and more challenging dataset collected with 12 outdoor cameras and 3 indoor cameras during 4 days. The training set contains 32,621 bounding boxes of 1,041 identities, and the testing set contains 93,820 bounding boxes of 3,060
identities.

In the experiments, the training data in source and target datasets are supposed to be labeled and unlabeled respectively. We employ the Rank-1 accuracy and mean Average Precision (mAP) \cite{zheng2015scalable} as evaluation metrics.  All the results are achieved under the single-query model without Re-Ranking \cite{Zhong_2017_CVPR} refinement for fair comparison.
\subsection{Implementation Details}
The parameters of ResNet-50~\cite{He2016DeepRL} are pre-trained on ImageNet, and other network parameters are all initialized randomly. The code is implemented on Pytorch and all images are resized to $384\times128$. Similar to~\cite{zhong2019invariance,SSG}, we perform random flipping, random cropping and random erasing~\cite{zhong2017random} for data augmentation in training.
The stochastic gradient descent with a momentum of 0.9 is adopted. At each iteration of Alg.~\ref{alg:1},  the learning rate is set to $ 1.5\times 10^{-4} $ for ResNet-50 base layers  and $3\times 10^{-5}$ for other layers in the first 20 epoches. The learning rate drops with 0.1 for every 60 epochs. The training at each iteration lasts for 260 epochs and  the mini-batch is composed of 32 images. During testing, the domain-shared features are used for matching.
\subsection{Comparisons with State-of-the-Art Methods }
The proposed work is compared with 18 state-of-the-arts methods under the same setting, including hand-crafted feature based approaches (LOMO~\cite{liao2015person}, BoW~\cite{zheng2015scalable} and UMDL~\cite{Peng2016UnsupervisedCT}), pseudo label based methods (CAMEL~\cite{yu2017cross}, PUL~\cite{fan2018unsupervised}, BUC~\cite{lin2019bottom},UDAP~\cite{song2018unsupervised}, SSG~\cite{SSG} and MAR~\cite{yu2019unsupervised}), GAN-based deep learning methods (SPGAN~\cite{Deng2017ImageImageDA}, PTGAN~\cite{Wei2017PersonTG}, CamStyle~\cite{yu2017cross}, HHL~\cite{HLL}, UCDA-CCE ~\cite{qi2019novel} and ECN~\cite{zhong2019invariance}) and other deep transfer learning approaches ( TJ-AIDL~\cite{wang2018transferable},  ARN~\cite{li2018adaptation} and EANet~\cite{wang2018transferable}).
\begin{table}[ht]
\begin{center}
\caption{Comparison with state-of-the-art methods on the Market-1501(Market) and DukeMTMC-reID(Duke) datasets.}\label{tab:M_and_D}
\resizebox{\columnwidth}{!}{
\begin{tabular}{l|c|cc|cc}
\hline
\multirow{2}{*}{ Methods}&\multirow{2}{*}{ Reference} &\multicolumn{2}{c}{
  Duke$\rightarrow$Market} &\multicolumn{2}{|c}{
  Market$\rightarrow$Duke}\\
\cline{3-6}
~&~&mAP&Rank-1&mAP&Rank-1\\
\hline
\hline
 LOMO&CVPR'15&8.0&27.2&4.8&12.3\\
 BoW&ICCV'15&14.8&35.8&8.3&17.1\\
 UMDL&CVPR'16&12.4&34.5&7.3&18.5\\
\hline
 CAMEL  &  ICCV'17& 26.3 & 54.5  	&   -   & -\\
 PUL  & ToMM'18     & 20.5 & 45.5 	&16.4	&30.0\\
 BUC & AAAI'19  &38.3 &66.2  &27.5 &47.4  \\
 MAR & CVPR'19      & 40.0 & 67.7    &48.0	&67.1 \\
 UDAP & arXiv'18     & 53.7 & 75.8  &49.0	&68.4 \\
 SSG & ICCV'19       &  58.3 &80.0  &53.4 &73.0  \\
\hline
 PTGAN& CVPR'18      & -    &38.6  	&-	    &27.4\\

 SPGAN & CVPR'18     &22.8  &51.5 	&22.3	&41.1\\
 {SPGAN+LMP}& CVPR'18  &26.7  &57.7 	&26.2	&46.4\\
 CamStyle& ICCV'17   &27.4  &58.8 	&25.1	&48.4\\
 HHL& ECCV'18        & 31.4 &62.2 	&27.2	&46.9\\
 UCDA-CCE&  ICCV'19    & 34.5 &64.3 & 36.7& 55.4    \\
 ECN & CVPR'19  & 43.0 & 75.1 &40.4	&63.3 \\
\hline
 TJ-AIDL& CVPR'18    & 26.5 & 58.2  &23.0	&44.3\\
 ARN &  CVPRW'18       & 39.4 &70.3  &33.4	&60.2\\
 EANet & arXiv'18    & 51.6  &78.0    &48.0	&67.7   \\
\hline
 \textbf{Ours} &  This work&\textbf{67.8}	&\textbf{86.4}	&\textbf{63.9}	&\textbf{77.6}	\\
\hline
\end{tabular}}
\end{center}
\end{table}

\begin{table}[t]
\footnotesize
\centering
\begin{center}
\caption{~~Performance gain analysis. D: Direct Transfer; U: Unsupervised Learning; G: Performance Gain. }\label{tab:gain}
\resizebox{\columnwidth}{!}{
\begin{tabular}{l|l|cc|cc}
\hline
\multirow{2}{*}{ Methods}&\multirow{2}{*}{ {~D/U/G}}&\multicolumn{2}{c}{
  ~Duke $\rightarrow$ Market~} &\multicolumn{2}{|c}{
  ~Market $\rightarrow$ Duke~}\\
\cline{3-6}
~&~~~~&~~ mAP~&~~ Rank-1~& mAP~&~~ Rank-1~\\

\hline
\hline
 \multirow{3}{*}{MAR}&~~ D&24.6&46.2~	&~28.8~&43.1~\\
\cline{2-6}
~&~~ U&40.0&67.7~	&~48.0~&67.1~\\
\cline{2-6}
~& G(U-D)&15.4&21.5~	&~19.2~&24.0~\\
\hline
 \multirow{3}{*}{ECN}&~~ D&17.7&43.1~	&~14.8~&28.9~\\
\cline{2-6}
~&~~ U&43.0&75.1~	&~40.4~&63.3~\\
\cline{2-6}
~& G(U-D)&25.3&32.0~	&~25.6~&34.4~\\
\hline
 \multirow{3}{*}{UDAP}&~~ D&19.1&46.8~	&~11.9~&27.3~\\
\cline{2-6}
~&~~ U&53.7&75.8~	&~49.0~&68.4~\\
\cline{2-6}
~& G(U-D)&34.6&29.0~	&~37.1~&41.1~\\
\hline
 \multirow{3}{*}{SSG}&~~ D& 26.6& 54.6 &16.1 &30.5 \\
\cline{2-6}
~&~~ U&  58.3 &80.0  &53.4 &73.0 \\
\cline{2-6}
~& G(U-D)&31.7&25.4~	&~37.3~&\textbf{42.5}~\\
\hline
 \multirow{3}{*}{Ours}&~~ D&23.4	&51.5		&22.9	&40.6\\
\cline{2-6}
~&~~ U&67.8	& 86.4	& 63.9	& 77.6	\\
\cline{2-6}
~& G(U-D)&\textbf{44.4}&\textbf{34.9}	&\textbf{41.0}	&37.0\\
\hline
\end{tabular}}
\end{center}
\end{table}
\begin{table}[ht]
\begin{center}
\caption{Comparison with state-of-the-art methods on the MSMT17 dataset.}\label{tab:MSMT}
\resizebox{\columnwidth}{!}{
\begin{tabular}{l|c|cc|cc}
\hline
\multirow{2}{*}{Methods}&\multirow{2}{*}{Reference} &\multicolumn{2}{c}{Market$\rightarrow$MSMT17} &\multicolumn{2}{|c}{Duke$\rightarrow$MSMT17}\\
\cline{3-6}
~&~&mAP~&Rank-1&mAP&Rank-1 \\

\hline
 PTGAN&CVPR'18&2.9&10.2&3.3&11.8\\
 ECN&CVPR'19& 8.5&25.3&10.2&30.2\\
 SSG&ICCV'19&13.2&31.6&13.3&32.2\\
\hline
\textbf{Ours}&This work&\textbf{20.8}&\textbf{44.5}&\textbf{21.6}&\textbf{46.7} \\
\hline
\end{tabular}}
\end{center}
\end{table}

The comparative results on Market-1501 and DukeMTMC-reID are shown in Table~\ref{tab:M_and_D}, and it is evident that:

\begin{table}[t]
\footnotesize
\begin{center}
\caption{Ablation studies of the proposed model and loss. IA denotes two individual attention modules are used to the domain-shared and domain-specific part.}\label{tab:daam_loss}
\resizebox{\columnwidth}{!}{
\begin{tabular}{l|cc|cc}
\hline
\multirow{2}{*}{Methods} &\multicolumn{2}{c}{
 {Duke $\rightarrow$
 Market}}&\multicolumn{2}{|c}{
  {Market $\rightarrow$
 Duke}} \\

\cline{2-5}
~ &  mAP& Rank-1& mAP& Rank-1 \\
\hline
\hline
	 {Supervised Learning}	&77.2	&90.8		&70.1	&84.1		\\
\hline
	 {Direct Transfer}	&23.4	&51.5		&22.9	&40.6		\\
\hline
     {Pseudo Label+Hard-Label} &61.1	&81.8	&56.2	&71.1	\\

    {Pseudo Label+LSR} &62.4	&82.7	&56.9	&72.4	\\
     {Pseudo Label+Soft-Label}	& \textbf{63.9}	& \textbf{83.7}	& \textbf{57.8}	& \textbf{73.3}	\\
\hline
     {Baseline+DSH+DSP}	&63.2	&82.3	&59.7	&74.0	\\
     {Baseline+DSH+DAAM}	&64.7	&84.4	&60.8	&74.9	\\
     {Baseline+DSH+DSP+IA}	&66.6	&85.2	&61.7	&76.0	\\
 {Baseline+DSH+DSP+DAAM}	&\textbf{67.8}	& \textbf{86.4}	& \textbf{63.9}	& \textbf{77.6}	\\
\hline
\end{tabular}}
\end{center}
\end{table}

\noindent(1) Our method outperforms hand-crafted feature based approaches(LOMO, BoW and UMDL) by a large margin, because the deep network model can learn more discriminative representations than hand-crafted features.

\noindent(2) The proposed method significantly exceeds  the pseudo label based unsupervised Re-ID models. In particular, we achieve mAP = 67.8\%(63.9\%) on Market-1501(DukeMTMC-reID), which outperforms the best unsupervised method SSG by +9.5\%(+10.5\%). A key reason is that pseudo labels are approximated and inaccurate, and the proposed method regards them as soft labels rather than hard labels.

\noindent(3) Compared with GAN-based methods, the proposed method can achieve higher performance without generating new images. It indicates that the proposed method can make use of the unlabeled data more effectively.

\noindent(4) Compared with ARN which also separates the feature to the DSH and DSP parts, our advantages are obvious. The reason is that the proposed attention can separate the feature map more directly than the encoder used in ARN.

Due to different baselines of the methods may effect the final performances, an additional experiment is conducted to better validate the effects of different methods. Specifically, we consider the performance of the direct transfer (D) of each method as the baseline, and evaluate the performances gain (G) between the final result (U) and the corresponding baseline. As shown in Table~\ref{tab:gain}, the proposed method can improve the baseline more significantly. Specifically, the previous best method UDAP improves mAP by +34.6\% (+37.1\%) on Market-1501 (DukeMTMC-reID) than the direct transfer, while the corresponding gain of the proposed method is +44.4\% (+41.0\%) on Market-1501 (DukeMTMC-reID).

In addition, we also evaluate the proposed method on a larger and challenging dataset MSMT17. As shown in Table~\ref{tab:MSMT}, the proposed method clearly exceeds  three existing  methods including PUL, ECN and SSG. Specifically, compared to the current best method SSG, we improve the performance by +12.9\%(+18.5\%) in Rank-1 accuracy when test on Market-1501 (DukeMTMC-reID).
It shows that our method still works well even with a larger and more complicated dataset.

\subsection{Ablation Study}
In this section, two groups of ablation studies are conducted to evaluate contributions of two proposed components, including the soft-label based loss Eq.~(\ref{eq:cro_t}) and the DAAM, as shown in Table~\ref{tab:daam_loss}.

\noindent\textbf{Effect of The Soft-label Based Loss.}
In the first group of experiments, the DAAM and the DSP branch are removed firstly and we aim to evaluate the individual effect of the loss. Specifically,  based on the estimated pseudo labels, 3 losses are adopted on the target data respectively,  including 1) ``Pseudo Label + Hard-Label'' which performs the hard-label based loss Eq.~(\ref{eq:cro_s}) on the pseudo labels directly (as same to \cite{fan2018unsupervised,song2018unsupervised}), 2) ``Pseudo Label + LSR'' where the LSR loss \cite{szegedy2016rethinking} is adopted and 3) ``Pseudo Label + Soft-Label'' corresponding to the proposed soft-label based loss. As shown in Table ~\ref{tab:daam_loss}, the proposed soft-label based loss outperforms others clearly.
The reason is that the pseudo labels are approximated and inaccurate, and the proposed loss can alleviate the over-fitting on the unreliable pseudo labels.
In addition, compared with LSR which fix the original distribution by the uniform distribution, the proposed soft-label based loss utilizes the relationship between the samples and  the groups, which is important to unsupervised learning.

\noindent\textbf{Effect of The DAAM.}
In this group of experiments, the ``Pseudo Label+Soft-Label'' is regarded as the baseline. Firstly, we add the DSP branch to the baseline without the DAAM (denoted as ``Baseline+DSH+DSP'') and with the DAAM (denoted as ``Baseline+DSH+DAAM'') respectively.
Then, two independent attention modules are introduced to the DSH and DSP branches independently (denoted as ``Baseline+DSH+DSP+IA''). In addition, the performances of  full model ``Baseline+DSH+DSP+DAAM'' are listed as reference. The comparison results in Table ~\ref{tab:daam_loss} demonstrates the effectiveness of the DAAM. In particular, the performance gain compared to  ``Baseline+DSH+DSP+IA'' indicates that the improvement of our method is caused by the residual mechanism which can separate the feature map into the DSH and DSP parts effectively, rather than just employing the attention techniques.

\begin{figure}[t]
\centering
\includegraphics[width=\columnwidth]{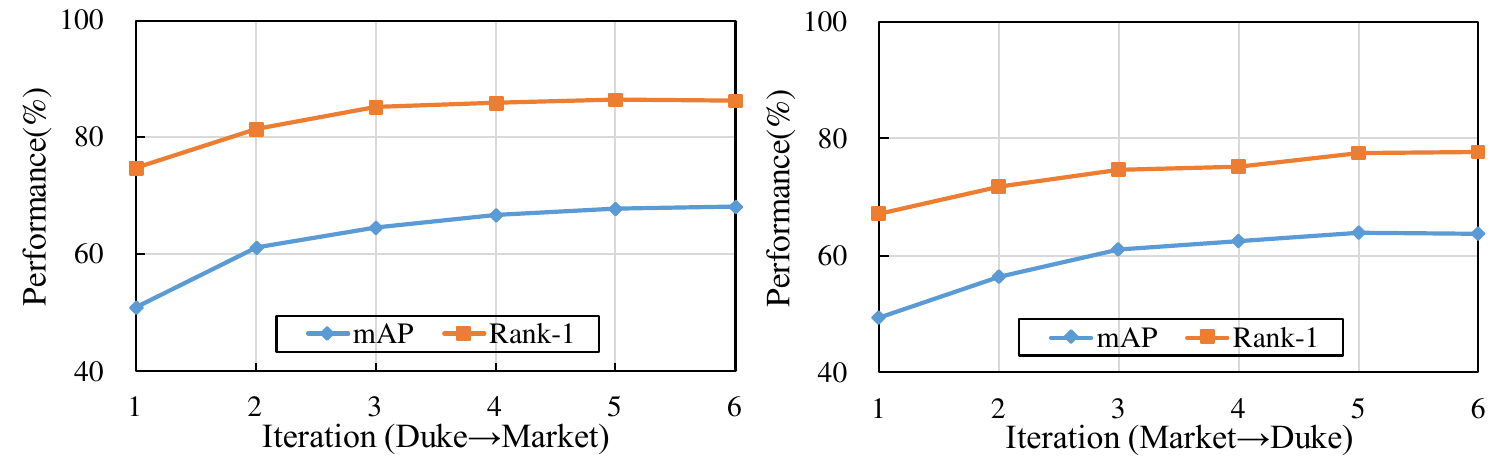}
\caption{~~Evaluation with different $Iteration$.} 
\label{fig:iteration}
\end{figure}
\begin{figure}[t]
\centering
\includegraphics[width=\columnwidth]{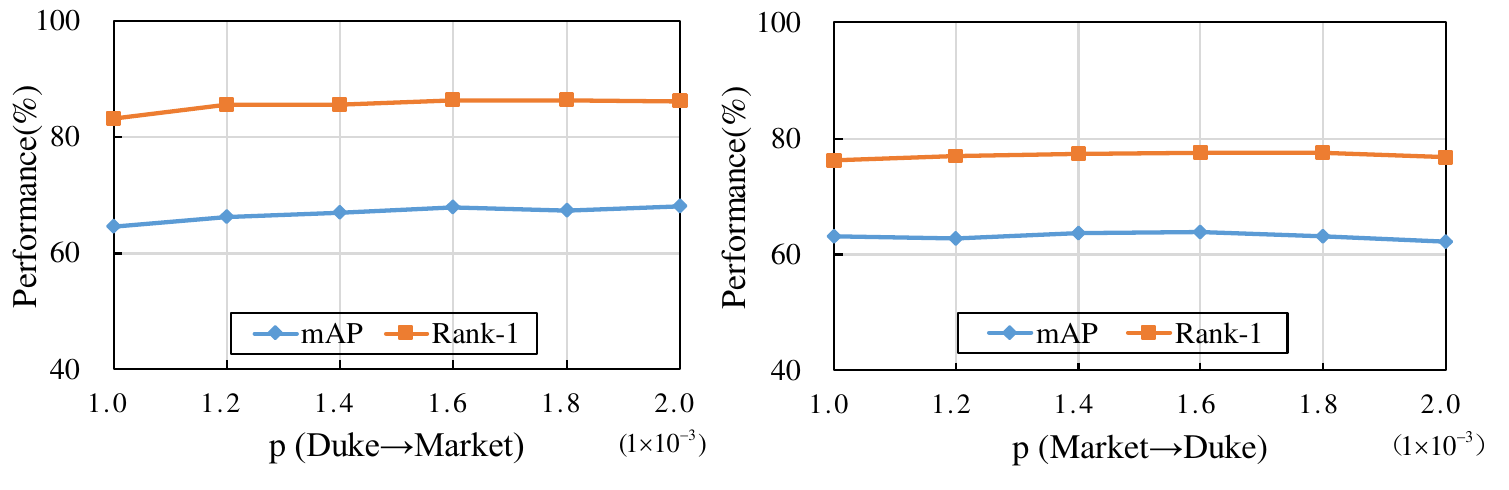}
\caption{~~Evaluation with different percentage $p$.}
\label{fig:p}
\end{figure}
\begin{figure}[t]
\centering
\includegraphics[width=\columnwidth]{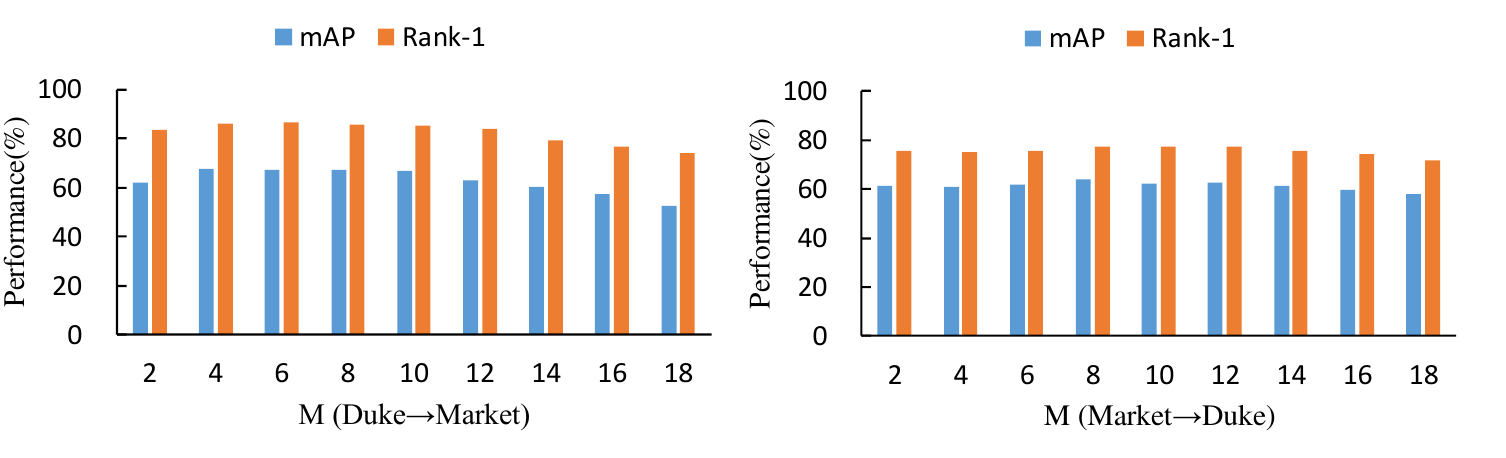}
\caption{~~Evaluation with different minimum number $M$ of a cluster.}
\label{fig:k}
\end{figure}

\subsection{Parameter Analysis}
In this section, we mainly evaluate the influence of two main hyper-parameters, including the number of iterations in in Alg.~\ref{alg:1} and the hyper-parameters related with DBSCAN.

\noindent\textbf{The Number of Iterations.} 
We evaluate the learned model after $Iterations=1,2,...,6$ in Alg.~\ref{alg:1} respectively, and the evaluation results are shown in Fig.~\ref{fig:iteration}. As the model becomes stronger in each iteration and more reliable pseudo labels of target images are generated, the performance is improved in early iterations. Finally, it converges after 5 iterations for both datasets.

\noindent\textbf{The Robustness of Parameters of DBSCAN.}
Here we evaluate the minimum size $M$ of a cluster and percentage $p$ for DBSCAN. (1)As shown in Fig.~\ref{fig:p}, We studied the effects of different proportions on the experimental results. It is observed that our approach is very stable and does not fluctuate greatly with different $p$. 
(2)We also set the minimum size of a cluster $M$ as $2, 4,...,18$ respectively and the results are shown in Fig.~\ref{fig:k}. We observe that our model learning is stable within a wide range for different $M$. Especially, when we transfer knowledge from Market-1501 to DukeMTMC-reID, the experimental result changes by no more than $1\%$ on Rank-1 accuracy as the change of $M$.


\section{Conclusion}

In the paper, we have proposed a novel unsupervised cross-domain transfer learning network architecture by using attention model for Re-ID task.
With the attention model based on residual mechanism, it can transfer knowledge from the labeled dataset to the unlabeled dataset by jointly modelling the domain-shared and domain-specific features.
Moreover, it differs significantly from existing methods in that a soft label loss is proposed to alleviate the negative effect of inaccuracy pseudo labels. Extensive experiments on Market-1501, DukeMTMC-reID and MSMT17 datasets have demonstrated the effectiveness and robustness of the proposed model.
\section*{Acknowledgments}

I am thankful to and fortunate enough to get support from the National Nature Science Foundation of China (NSFC) under Grants 61972030 and 61702515. Also, I would like to extend our sincere esteems to all staff in laboratory for their timely support.

{\small
\bibliographystyle{aaai}
\bibliography{7576}
}

\end{document}